\documentclass{article}
\usepackage{ijcai24}
\usepackage{subfigure}
\usepackage{times}
\usepackage{soul}
\usepackage{url}
\usepackage[hidelinks]{hyperref}
\usepackage[utf8]{inputenc}
\usepackage[small]{caption}
\usepackage{graphicx}
\usepackage{amsmath}
\usepackage{amsthm}
\usepackage{booktabs}
\usepackage{algorithm}
\usepackage{algorithmic}
\usepackage[switch]{lineno}
\usepackage{color}
\usepackage{makecell}
\title{3DBench: A Scalable 3D Benchmark and Instruction-Tuning Dataset}

\author{
Junjie Zhang$^1$
\and
Tianci Hu$^1$
\and
Xiaoshui Huang$^2$
\and
Dan Zeng$^1$
\and
Yongshun Gong$^3$
\\
\affiliations
$^1$Shanghai University\\
$^2$Shanghai AI Laboratory\\
$^3$Shandong University\\
\emails
\{junjie\_zhang, yinsh\}@shu.edu.cn,
huangxiaoshui@pjlab.org.cn,
}

\begin{document}

\maketitle

\begin{abstract}
   % Large language models have demonstrated significant potential in the point cloud field, expanding their capabilities to incorporate an additional modality. 
Evaluating the performance of Multi-modal Large Language Models (MLLMs), integrating both point cloud and language, presents significant challenges. The lack of a comprehensive assessment hampers determining whether these models truly represent advancements, thereby impeding further progress in the field. Current evaluations heavily rely on classification and caption tasks, falling short in providing a thorough assessment of MLLMs. A pressing need exists for a more sophisticated evaluation method capable of thoroughly analyzing the spatial understanding and expressive capabilities of these models. To address these issues, we introduce a scalable 3D benchmark, accompanied by a large-scale instruction-tuning dataset known as 3DBench, providing an extensible platform for a comprehensive evaluation of MLLMs. Specifically, we establish the benchmark that spans a wide range of spatial and semantic scales, from object-level to scene-level, addressing both perception and planning tasks. Furthermore, we present a rigorous pipeline for automatically constructing scalable 3D instruction-tuning datasets, covering 10 diverse multi-modal tasks with more than 0.23 million QA pairs generated in total. Thorough experiments evaluating trending MLLMs, comparisons against existing datasets, and variations of training protocols demonstrate the superiority of 3DBench, offering valuable insights into current limitations and potential research directions.
\end{abstract}

\section{Introduction}

Recently, there have been significant advancements in multi-modal large language models (MLLMs) \cite{li2023blip,zhu2023minigpt,radford2021learning}, catalyzing a profound revolution across various tasks. Large-scale instruction-tuning data is essential to harness the capabilities of MLLMs. To facilitate the integration of large models into the 3D domain, our goal is to establish a scalable evaluation benchmark specifically designed for assessing 3D-LLMs. Additionally, we elaborate on the detailed development of a large-scale dataset to address the scarcity of instruction-tuning datasets in the 3D domain.

\begin{figure}
\centering
\includegraphics[width=0.8\linewidth]{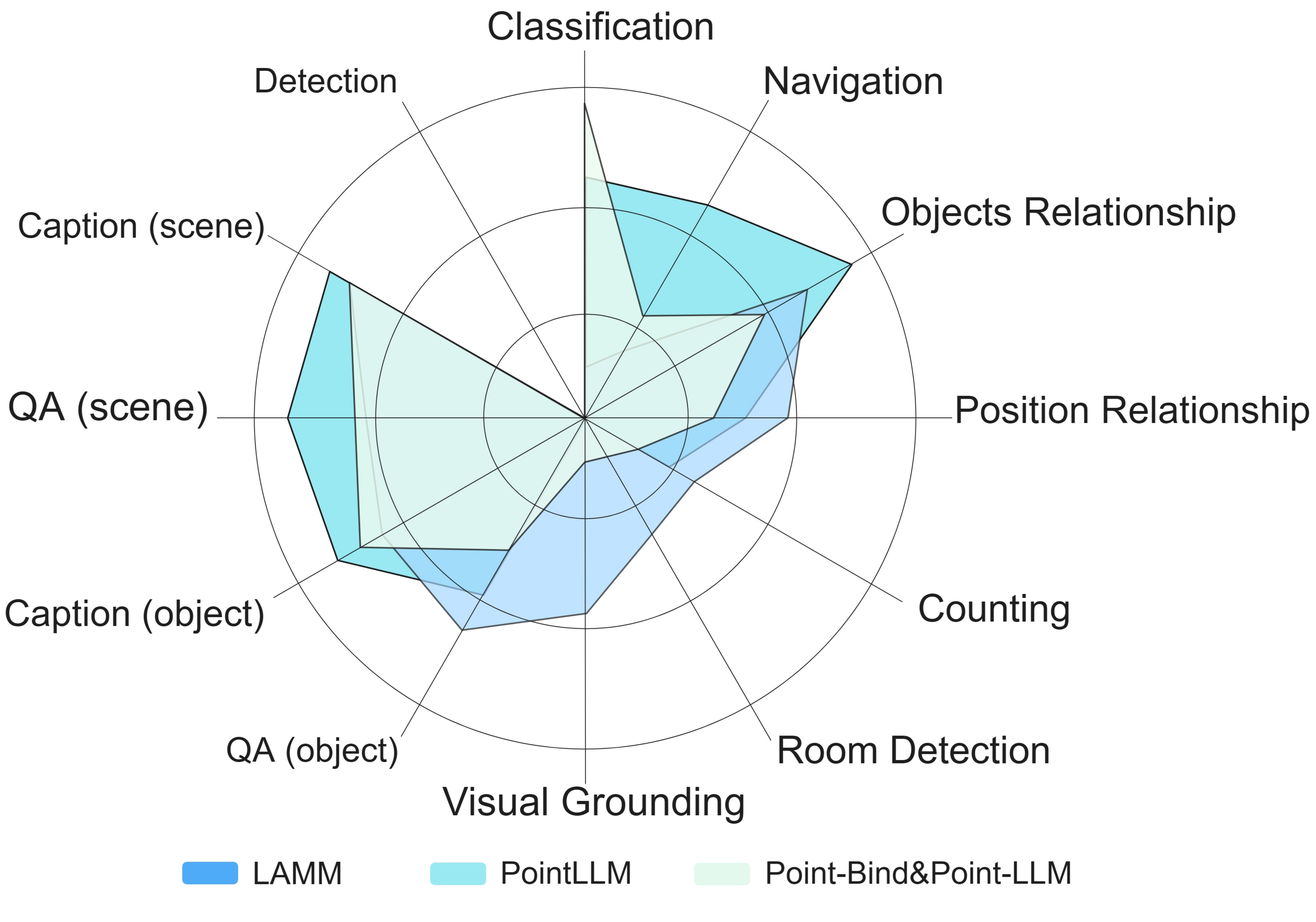}
\caption{Zero-shot evaluation of three state-of-the-art 3D-LLMs on proposed 3DBench with ten multi-modal tasks.}
\label{figure:chart}
\end{figure}

Existing instruction-tuning datasets for 3D-LLMs originate from publicly open datasets. PointLLM \cite{xu2023pointllm} utilizes Cap3D \cite{luo2023scalable}, a 3D object captioning dataset derived from Objaverse \cite{deitke2023objaverse}. Point-Bind \& Point-LLM \cite{guo2023point} is trained using Ulip \cite{xue2023ulip}, constructed based on ShapeNet \cite{chang2015shapenet}. LlaVa \cite{wang2023vigc} efficiently incorporates the image modality into large models by prompting GPT-4 \cite{fu2022does} with captions from the COCO dataset \cite{lin2014microsoft}. They also conduct experiments to validate the effectiveness of using GPT-generated \cite{openai2023gpt} image-text pairs as training data.

However, existing \textbf{benchmarking} efforts primarily focus on single-object classification and captioning, neglecting an effective evaluation of the spatial understanding capabilities of large models for complex and scene-oriented point clouds. Benchmarks for point cloud scenes, such as Visual Grounding (VG), Detection, and VQA tasks outlined by LAMM \cite{yin2023lamm}, can assess a model's interpretation of object positions in the given scene and its general knowledge inferred from language model. However, these tasks still do not sufficiently address the comprehension of multi-object relationships and the room-area perception capabilities, commonly referred as the scene reasoning ability of large models.

Additionally, high-quality \textbf{instruction-tuning datasets} in the 3D domain remain limited \cite{Wei_Bosma_Zhao_Guu_Yu_Lester_Du_Dai_Le_2021}. Current datasets are mainly collected to introduce large models and are generated based on open 3D datasets. This approach introduces a potential risk of data leakage in pre-training models \cite{dai2024instructblip}. Significant distinctiveness in task categories exists among datasets, with few benchmarks for each one and a lack of uniformity, making it challenging to accurately and comprehensively assess the capabilities of 3D large models.

\begin{figure}[t]
\centering
\includegraphics[width=0.7\linewidth]{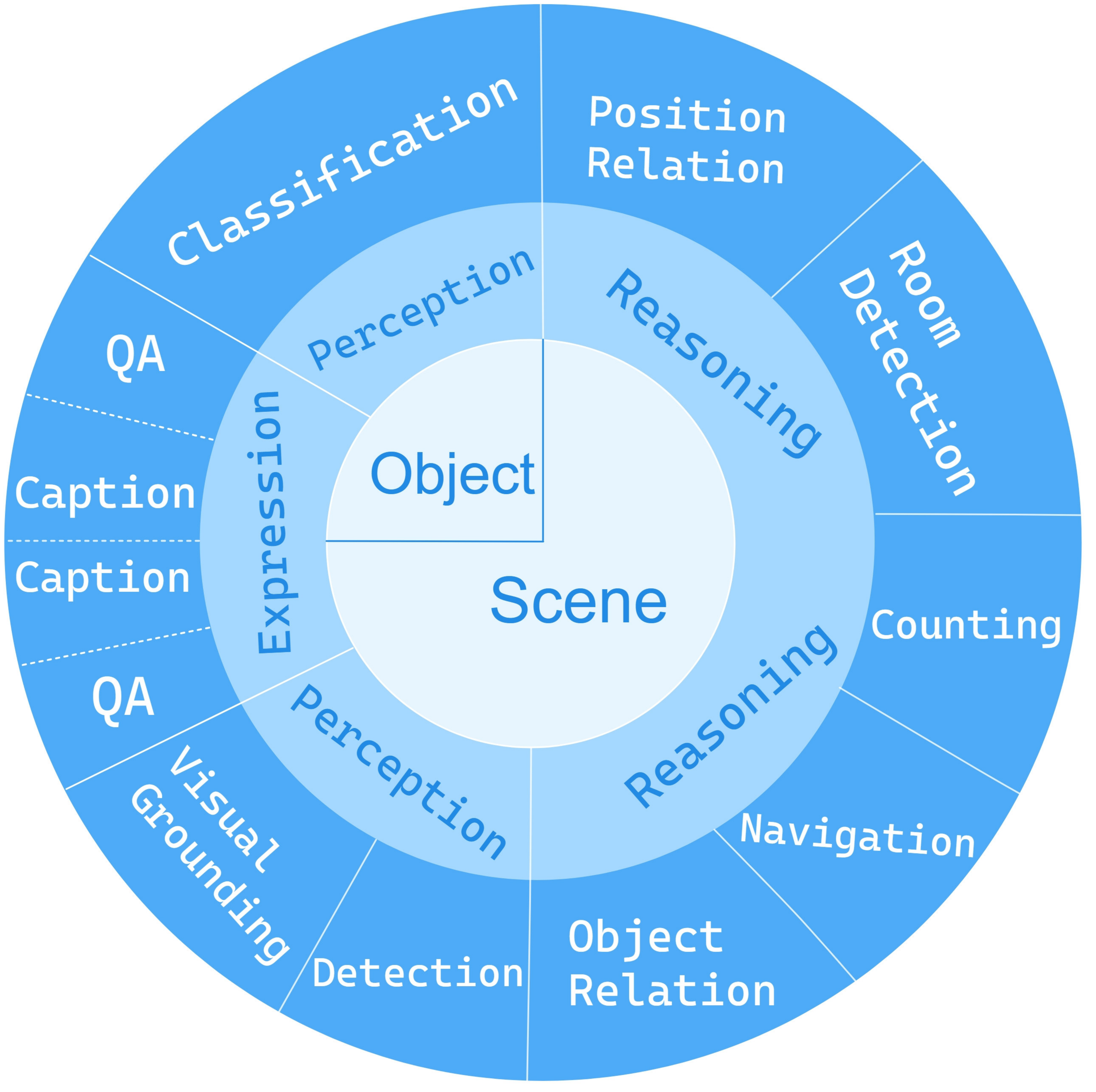}
\caption{
The current task overview of 3DBench. 3DBench comprehensively addresses the complexity of both spatial and logical aspects, categorizing ten individual tasks into three levels. Future tasks can seamlessly integrate into this framework.
}
\label{fig:overview}
\end{figure}

To address above issues, We propose a scalable benchmark comprising ten multi-modal tasks and three evaluation metrics. As shown in Fig. \ref{fig:overview}, the tasks include common ones such as classification, VG, detection, and counting. Building on this foundation, we extend VG to include a room detection task. Additionally, we expand object understanding from 2D to 3D, requiring reasoning about the relative positions and attribute relationships of multiple objects. To evaluate the quality of large model dialogues and long-text generation capabilities, we include QA and caption tasks. Finally, we design a navigation benchmark to evaluate the spatial planning capabilities of 3D-LLMs. Evaluation metrics consist of three types: (1) For tasks with textual outputs, like relationship reasoning, we adopt a heuristic approach to instruct ChatGPT to score the predictions. (2) Traditional metrics such as precision and mAP are employed to evaluate detection and VG tasks. Furthermore, we introduce a path loss to assess navigation task. (3) For questions with simple-structured answers like classification and counting, we generate 300 to 2000 multiple-choice questions \cite{liu2023mmbench} and calculate the accuracy of output options.

The process of acquiring a large-scale instruction-tuning dataset consists of two main steps. During the initial step, we extract comprehensive metadata from the Procthor simulation framework \cite{deitke2022️}. This metadata comprises depth maps and corresponding diverse types of ground-truth for both objects and scenes. Utilizing the depth maps, we manage to reconstruct a variety of 3D objects and scenes. In the second step, we utilize the ground-truth to inspire GPT to generate knowledge for textual tasks. By incorporating the ground-truth into diverse dialogue templates, we obtain a instruction-tuning dataset for various fine-grained tasks. 
In summary, our contributions can be summarized in three main aspects:

\begin{itemize}
\item \textbf{Evaluation Benchmark:} We develop a benchmark that spans from the object to the scene scale, covering dimensions of perceptual and reasoning abilities. It includes ten diverse multi-modal tasks, assessed with three types of customized evaluation metrics. The benchmark is enhanced and extended beyond existing standards through the incorporation of our proposed tasks and metrics.

\item \textbf{Instruction-tuning Dataset:} We design an approach for the automatic acquisition of a large-scale 3D instruction-tuning dataset, resulting in 34,000 point clouds of everyday objects and 30,000 indoor scenes containing them. The dataset comprises over 0.23 million QA pairs, encompassing ten fine-grained tasks.

\item \textbf{Experiments and Observations:} Experimental validation substantiates the efficacy of our proposed dataset. As shown in Fig. \ref{figure:chart}, we conduct a thorough quantitative evaluation of existing 3D-LLMs using 3DBench. Observations cover the quality of the generated dataset and variations in training protocols, offering insights for future explorations.

\end{itemize}

\begin{figure*}[t]
	\centerline{
		\includegraphics[width=1\linewidth]{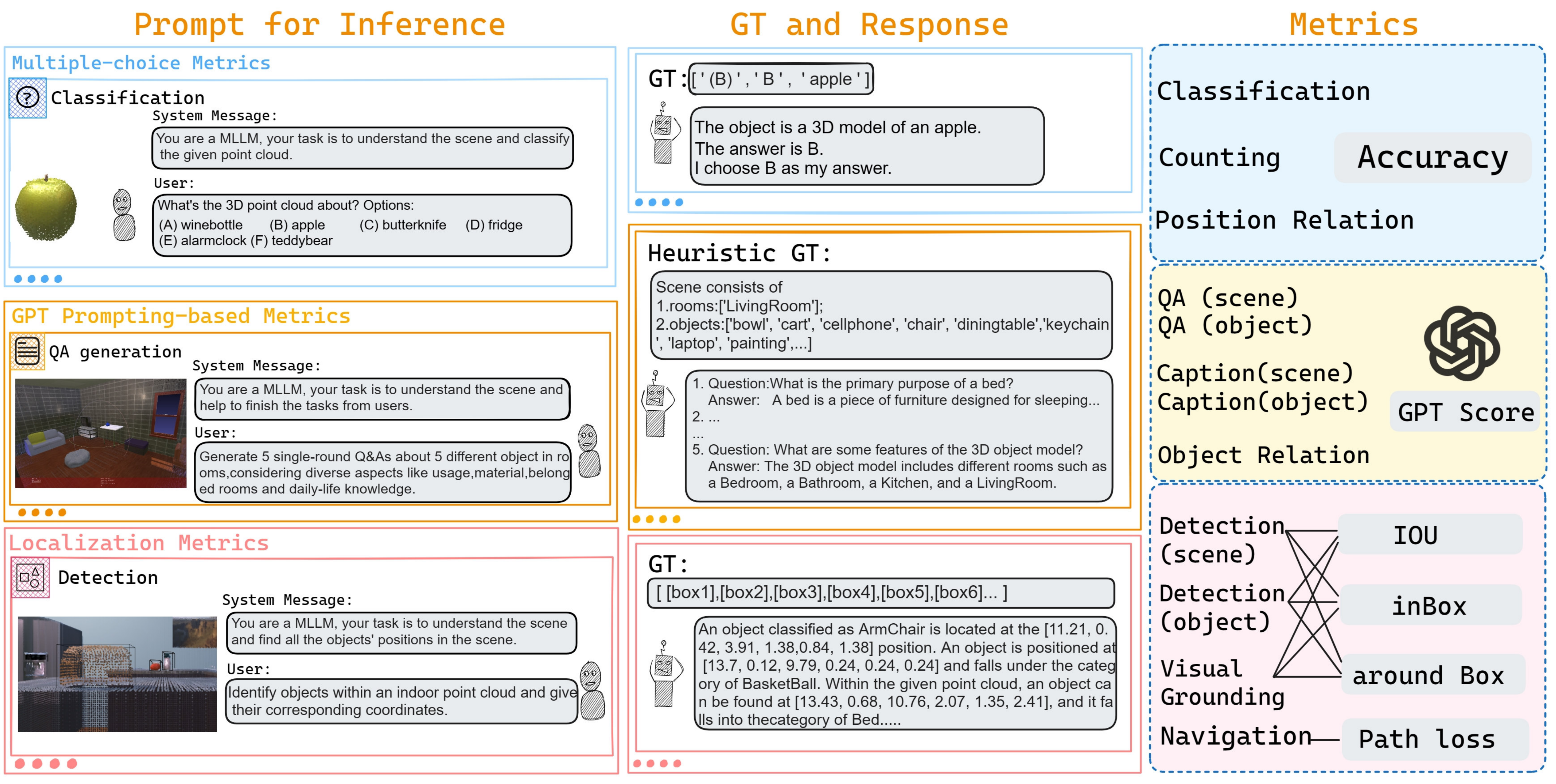} 	
	}
	\caption{Overview of the 3DBench benchmark, encompassing ten 3D computer vision tasks and metrics from three perspectives, including traditional accuracy, IOU metric, GPT scores, and the novel path loss metric introduced by us.
}
	\label{fig:evaluation}
\end{figure*}

\section{Related Work}

% 3D large models demonstrate excellent zero-shot capabilities in handling multimodal downstream tasks \cite{hao2022language} and have achieved promising qualitative results in existing work. Nevertheless, the absence of a thorough quantitative assessment for large 3D models impedes their progression in enhancing proficiency in handling 3D tasks.

The proposed 3DBench comprises two main contributions: benchmark and dataset. To assess the distinctiveness of existing work and our contributions, we review related works from two perspectives.

\subsection{Multi-modal Evaluation Benchmark}
Several existing benchmarks evaluate MLLMs across a range of tasks \cite{hao2022language}. MMBench~\cite{liu2023mmbench}, with a focus on perception and reasoning, designs 20 fine-grained tasks that cover the three-tier capacity dimensions of MLLMs. MME \cite{fu2023mme} curates all instruction-answer pairs manually to prevent data leakage and utilizes 14 sub-tasks to assess the perceptual and cognitive capabilities of VLLMs. LAMM \cite{yin2023lamm} extends research to incorporate point clouds, creating three visual task benchmarks for 3D tasks to facilitate the assessment of scene-level perceptual abilities. To assess the text quality of large model outputs, AlpacaFarm \cite{dubois2023alpacafarm} introduces a swift and reliable GPT-4-based automated benchmark. GPT is commonly used to align target model outputs with answers, enhancing evaluation robustness. Traditional metrics such as Bleu \cite{Papineni_Roukos_Ward_Zhu_2001}, ROUGE \cite{Lin_2004}, and METEOR \cite{Banerjee_Lavie_2005} produce ineffective evaluation results due to answer bias, unlike GPT.
Although current benchmarks have made noteworthy contributions to evaluating MLLMs, there remains a deficiency in comprehensive quantitative assessment for 3D-LLMs.

\subsection{Instruction-tuning Dataset}
Instruction-tuning datasets consist of paired samples with instructions and their corresponding answers. Public multi-modal datasets such as 2D COCO Caption and 3D ShapeNet offer objects and respective captions. LlaVa prompts GPT to generate detailed image descriptions with concise subtitles and object positions. Subsequently, it introduces new modalities into LLMs through textual information associated with the images. LAMM goes a step further by embedding ground truth from traditional computer vision tasks into dialogue templates, incorporating rich metadata from the dataset into instruction fine-tuning data to enhance the MLLM's ability to handle diverse tasks. In the 3D domain, ULIP-2\cite{xue2023ulip} and Cap3D prompt MLLMs to automatically generate captions for images rendered from point clouds. Point-LLM\cite{guo2023point} uses prompts from ChatCaptioner to generate high-quality Objaverse descriptions, while PointLLM\cite{xu2023pointllm} leverages captions generated by Cap3D to prompt GPT for more detailed point cloud descriptions in a conversational format. However, existing 3D instruction-tuning datasets are relatively scarce, and their sources are often limited to publicly available datasets. The availability of high-quality instruction-tuning datasets for 3D tasks is indeed a critical need.

\section{3DBench}

% This chapter is structured as follows: in section 3.1, we delve into the design of 3DBench's multi-level evaluation system. Section 3.2 highlights the enhancements and additions made to existing tasks. Finally, in Section 3.3, we explain how we obtained a unified instruction-tuning dataset for scenes, object point clouds, and various fine-grained tasks.
\subsection{Overview}
% Humans can effortlessly engage in simple classification and description tasks for individual point clouds, but once such tasks are extended to the scene level, the time required increases significantly. As problems become more complex, involving multiple objects, the process of logical reasoning and comparative analysis comes into play. Tasks ascend from simple perception to spatial and logically more intricate levels of inference, representing real-world challenges.

3DBench, as aillustrated in Fig. \ref{fig:overview} and \ref{fig:evaluation}, stands out from existing 3D benchmarks for two key reasons. Firstly, 3DBench includes ten diverse tasks to thoroughly assess the capabilities of 3D models in addressing aforementioned challenges. Secondly, 3DBench is built upon a unified instruction-tuning dataset that is extensive and hierarchically rich.
It is grounded in human cognition of real-world tasks, illustrating a three-tiered assessment. The top-level dimension, denoted as L-1, signifies the most apparent object-scene scale perception dimension. Beyond this, we consider the characteristics of large models and the logical reasoning process, allowing us to categorize tasks into three distinct categories: perception, reasoning, and expression. This introduces the L-2 competency dimension, covering three dimensions: 1) single or multiple-instance perception; 2) multi-instance relational reasoning; 3) conversational and descriptive abilities. Additionally, we derive the L-3 tasks dimension from the L-2 competency dimension. The benchmark includes ten evaluation tasks and three types of evaluation metrics.

% We collected and carefully selected a range of existing tasks to identify the most suitable ones for MLLMs. 
% Additionally, we expanded object-level tasks to encompass scene-level tasks, introducing advanced challenges such as Room Detection, Position Relation, Object Relation, and Navigation. Moreover, we devised two text generation tasks to evaluate the expressive capabilities of MLLMs. To accommodate the diverse output formats of these tasks, we developed and summarized three evaluation methods: GPT, diverse metrics and multi-choices.

\begin{figure*}[t]
	\centerline{
		\includegraphics[width=0.9\linewidth]{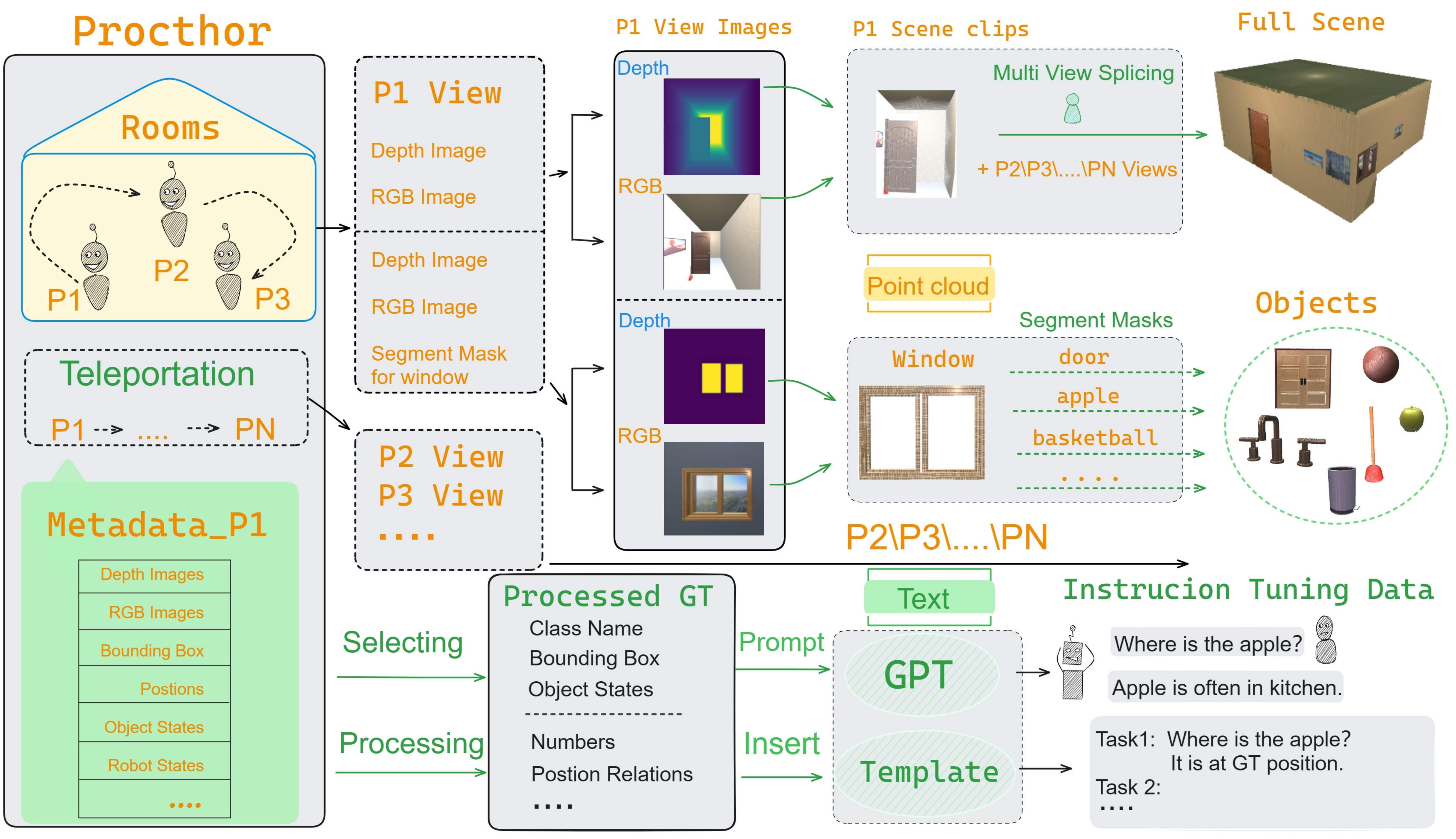} 	
	}
	\caption{Pipeline 3for generating the dataset. We are able to automatically collect instruction-tuning data for all detailed tasks in 3DBench. Each data sample comprises a point cloud (scene or object) along with the corresponding task dialogue.}
	\label{fig:pipeline}
\end{figure*}

\subsection{Evaluation Tasks}

% 1) heuristic GPT prompting score method.

% 2) Scene-based evaluation strategy

% 3) Navigation-based evaluation strategy

% \textbf{Existing Tasks.} We extensively collected tasks in the 3D domain and language models. After excluding tasks that existing large models cannot handle, we balanced and selected four tasks for our three-tier evaluation system: Classification, Visual Grounding, Detection, and Counting tasks.

\textbf{Expansion Tasks.} We systematically compile tasks from the 3D domain and language models. After excluding tasks beyond the capability of existing large models, we choose four tasks for our evaluation: classification, VG, detection, and counting. Building on prior works, we extend the scope of these existing tasks. Additionally, we expand the detection task to encompass the scene scale, instructing the model to predict bounding boxes for all rooms to assess the large model's overall perception capability in complex 3D scenes. Recognizing the expressive capacity of 3D-LLMs, we also introduce two textual generation tasks: generating detailed descriptions for long texts and generating dialogues for chat.

\textbf{Novel Tasks.} Drawing inspiration from 2D multi-instance tasks, we introduce tasks that require considering relationships and positional reasoning among multiple objects in real-world scenes, specifically, the multi-instance task within a given scene. Additionally, anticipating the capabilities of future large models in scene navigation and planning, we establish a navigation benchmark to assess the intricate scene perception and path planning abilities of 3D models.

\subsection{Evaluation Metrics}
\textbf{GPT Prompting-based Metrics.} As depicted in Fig. \ref{fig:evaluation}, we present a heuristic prompt tailored to address text generation tasks spanning different scales. When utilizing LLMs as the text evaluator, we provide GPT with information about all actual objects in the scene to avoid inflated scores caused by illusions. Moreover, this approach enables GPT to assess the authenticity of subtitles without depending solely on manual annotations. Furthermore, manual annotations may not necessarily represent the only correct solution for text generation. For evaluating text generation quality aligned with human habits, we recommend employing GPT for increased robustness.

\begin{figure}[t]
	\centerline{
		\includegraphics[width=\linewidth]{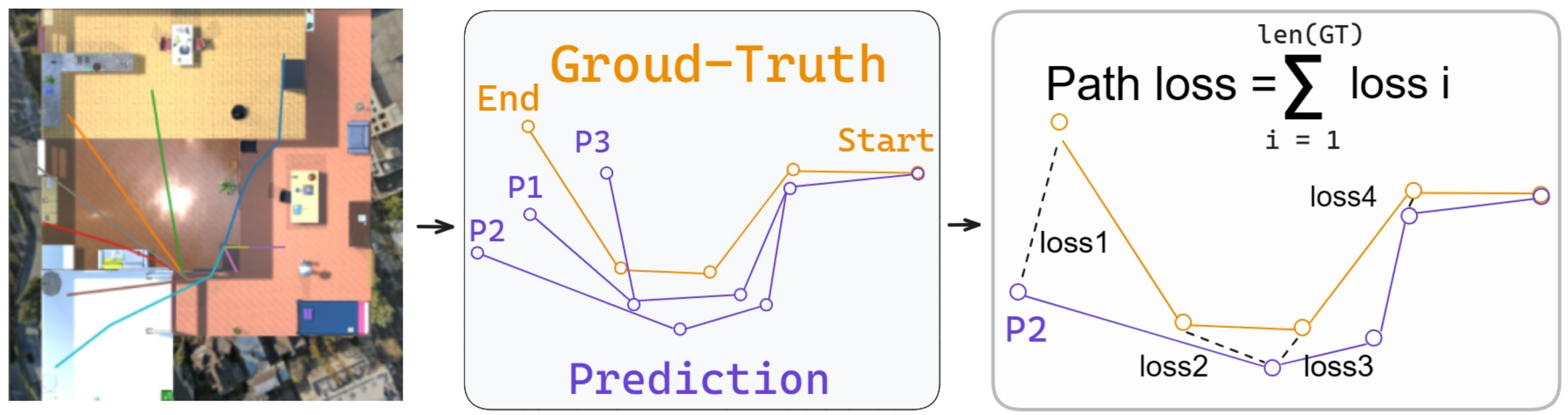} 	
	}

	\caption{The illustration of path loss. It is the distance accumulation of each endpoint on the longer trajectory (between GT and prediction) and its nearest neighbors on the other one.
}
	\label{fig:evaluationresults1}
\end{figure}

\textbf{Localization Metrics.} We replace the conventional IOU metric, known for yielding low scores in detection and visual grounding (VG) tasks, with an innovative approach. We evaluate whether the predicted object's center falls within the bounding box, resulting in the `in box' metric. Furthermore, we relax the criteria, permitting the predicted center to be within a one-meter radius of the ground-truth center, leading to the `around box' metric. Additionally, we propose a novel path loss metric for the navigation task as shown in Fig. \ref{fig:evaluationresults1}. It selects the longer trajectory between the prediction and ground-truth, measuring the distance between each endpoint of this trajectory and its nearest neighbor from the other one. The accumulated distance is compared against a pre-defined threshold to determine the success of the navigation.

\textbf{Multiple-choice Metrics.} 
We generate numerous multiple-choice questions for tasks with straightforward ground truth, encompassing classification, counting, and spatial relationships. To improve the ability of weaker 3D-LLMs to choose the correct answer, we direct all models to consider the correct answer as the only option. If a question is difficult to answer, we recommend all models to generate a random answer to avoid nonsensical responses.

\subsection{Dataset Construction}

\textbf{Pipeline.} The objective is to create large-scale 3D instruction-tuning datasets. To accomplish this, we design a scalable data construction pipeline. The pipeline comprises two main steps, as shown in Figure \ref{fig:pipeline}. In the initial step, we collect data for 30,000 houses using Procthor\cite{deitke2022️} and extract depth images and metadata ground-truth from the embodied AI simulation framework Ai2thor\cite{ai2thor}. In the second step, we reconstruct point clouds and derive instruction-tuning datasets for all tasks based on the ground-truth.

Specifically, we reconstruct scene point clouds from color and depth images of the scene and objects with instance segmentation. Prothor can automatically generate random complete indoor scenes and is built on a simulation robot training framework, enabling us to easily obtain metadata from diverse perspectives. Depending on the scene's size, we teleport the robot to different positions to capture complete scene images. Through this automated process, we acquire 30,000 completely random indoor scene point clouds and over 34,000 object point clouds representing 93 categories of everyday items. We use ground-truth to prompt GPT to acquire rich world knowledge, serving as our training data for tasks such as object relation, QA, and caption. To enhance the ability to handle diverse visual tasks, we process the original ground-truth to obtain results for tasks like counting and room detection. Additionally, we generate diverse dialogue templates tailored for different tasks, embedding the results into conversations to create an augmented instruction-tuning dataset.

\textbf{Data Statistics.} 
We construct a dataset of more than 0.23 million instruction-tuning samples, utilizing 224,000 for training and 8,000 for evaluation. For the training sets related to visual tasks, we extract over 20 samples from each scene, ensuring that the fine-tuning sample size for each task exceeds 10,000. For text generation tasks, we provide approximately 1,000 training samples for fine-tuning models. The distribution of our instruction-tuning dataset is depicted in Table \ref{tab:data}.

\begin{table}[t]
    \centering
    \small
    \begin{tabular}{c|ccc}
        \toprule
        Task &Train  & Test  & All \\
        \midrule
        Detection     & 30k          & 50     & 30k  \\
        QA(scene)  & 450          & 10     & 460  \\
        Caption(scene) & 450          & 10   & 460   \\
        Classification  & 33k          & 696    & 34k  \\
        Visual Grounding     & 60k      & 835  & 60k  \\
        Counting     & 15k          & 300   & 15k  \\
        Room Detection     & 30k          & 10   & 30k   \\
        Position Relationship     & 5k          & 185   & 1k   \\
        Objects Relationship     & 5k       & 10  & 10k  \\
        Navigation                   & 45k       & 6.7k    & 52k  \\
        \midrule
        Ten tasks             & \textbf{223k}     & \textbf{8k}              & \textbf{231k}          \\
        \bottomrule
    \end{tabular}
    \caption{Statistics on the distribution of 3DBench Dataset.}
    \label{tab:data}
\end{table}

\section{Experiments}

% In Section 4.1, we introduced the design of four comparative experiments. Section 4.2 outlined the experimental settings for all point cloud large models.

\subsection{Experiment Settings}
The experimental settings are detailed in Table \ref{tab:experiment settings}. Five groups of validation experiments are conducted to assess the effectiveness of our benchmark and dataset. As LAMM establishes benchmarks for detection, VQ, and VQA tasks, we use it as a robust baseline. It is originally trained on ShapeNet and 3RScan datasets, and then evaluated on the ScanNet dataset. 
Our experiments consist of five groups:
\begin{itemize}
    \item E1 \& E2: We initially assess the zero-shot performance of the LAMM model on the 3DBench test split (referred to as E1). Subsequently, employing the LAMM training framework, we exclusively train a model on our training split and evaluate its zero-shot performance on LAMM-Bench (noted as E2). The cross-validation results of three LAMM tasks (detection, VG, and VQA) for E1 \& E2 are compared to analyze differences between our dataset and publicly available ones.

    \item E3: We systematically categorize the 3DBench dataset into various scales and employ the same framework to train multiple LAMM models. The objective is to investigate how the expanded dataset scale influences the enhancement of LLM performance across three tasks: detection, VG, and classification.

    \item E4: We conduct a re-training of the LAMM model using the complete version of our 3DBench. Subsequently, we compare the performance of the re-trained model with its original version (assessing zero-shot ability) on our test split, encompassing ten tasks.

    \item E5: We assess the zero-shot performance of two additional 3D-LLMs using the complete version of 3DBench to evaluate their capabilities across ten tasks.
\end{itemize}

\begin{table}[t]
    \centering
    \small
    \resizebox {\linewidth} {!}{
    \begin{tabular}{c|cccc}
        \toprule
        Group Idx    & Model  & Training Split  & Test Split & Task \\
        \midrule
         E1  & LAMM    & ShapeNet \& 3RScan  &3DBench &Detection ; VG ; VQA  \\
         E2  & LAMM    & 3DBench  &ScanNet &Detection ; VG ; VQA  \\
        E3  & LAMM    & Variations of 3DBench  &3DBench  &Detection ; VG ; Classification \\
        E4  & LAMM    & 3DBench  &3DBench  & Ten Tasks \\
        E5  & All Three    &None  &3DBench  & Ten Tasks \\
        \bottomrule
    \end{tabular}}
    \caption{Details of five groups of experiment settings.}
    \label{tab:experiment settings}
\end{table}

% It is important to note that, due to cost considerations, not all tasks have to be trained on all scenes. For object recognition tasks, we did not train on all objects present in the 30k scenes. We believe that the 34k objects included in the initial five hundred randomly generated scenes are sufficient to cover all indoor objects.

% \subsection{Details of Evaluation.}

LAMM Settings: To ensure a fair comparison of the point cloud understanding abilities among three models, we conduct tests using the 7B versions for all models. During the re-training experiment with LAMM, we identify biases in various evaluation metrics related to output text lengths. As a result, we adjust the target length for different tasks, aiming to reveal the optimal performance of each model on the respective dataset.

PointLLM \& Point-LLM Settings: We maintain the default model parameter settings for both models. Following their guidelines, we uniformly sample point clouds to a fixed quantity. The evaluation encompass all ten tasks in 3DBench for PointLLM and Point-LLM. Due to the limitation of training data obtained from public sources, which is confined to the object level, and the potential vulnerability of inference results from untrained tasks to illusions, we focus on the evaluation of scene-level tasks. This aims to scrutinize the performance of large models when confronted with unfamiliar tasks and data.

\subsection{Cross-set Validation (E1 \& E2)}

% \begin{figure}[htbp]
% \centering
% 	% \begin{minipage}{0.49\linewidth}
%     \centering
%     \subfigure[]{
%         \includegraphics[width=0.22\textwidth]{result1_BMP.PDF}
%         \label{fig:subfig1}
%     }
%     \subfigure[]{
%         \includegraphics[width=0.23\textwidth]{result2_BMP.PDF}
%         \label{fig:subfig2}
%     }
%     \caption{(a) Zero-shot results on each other's benchmarks for the two datasets. (b) Results on the impact of dataset scale on LLM performance.}
%     \label{fig:main}
% \end{figure}

\begin{figure}[t]
	\centerline{
		\includegraphics[width=\linewidth]{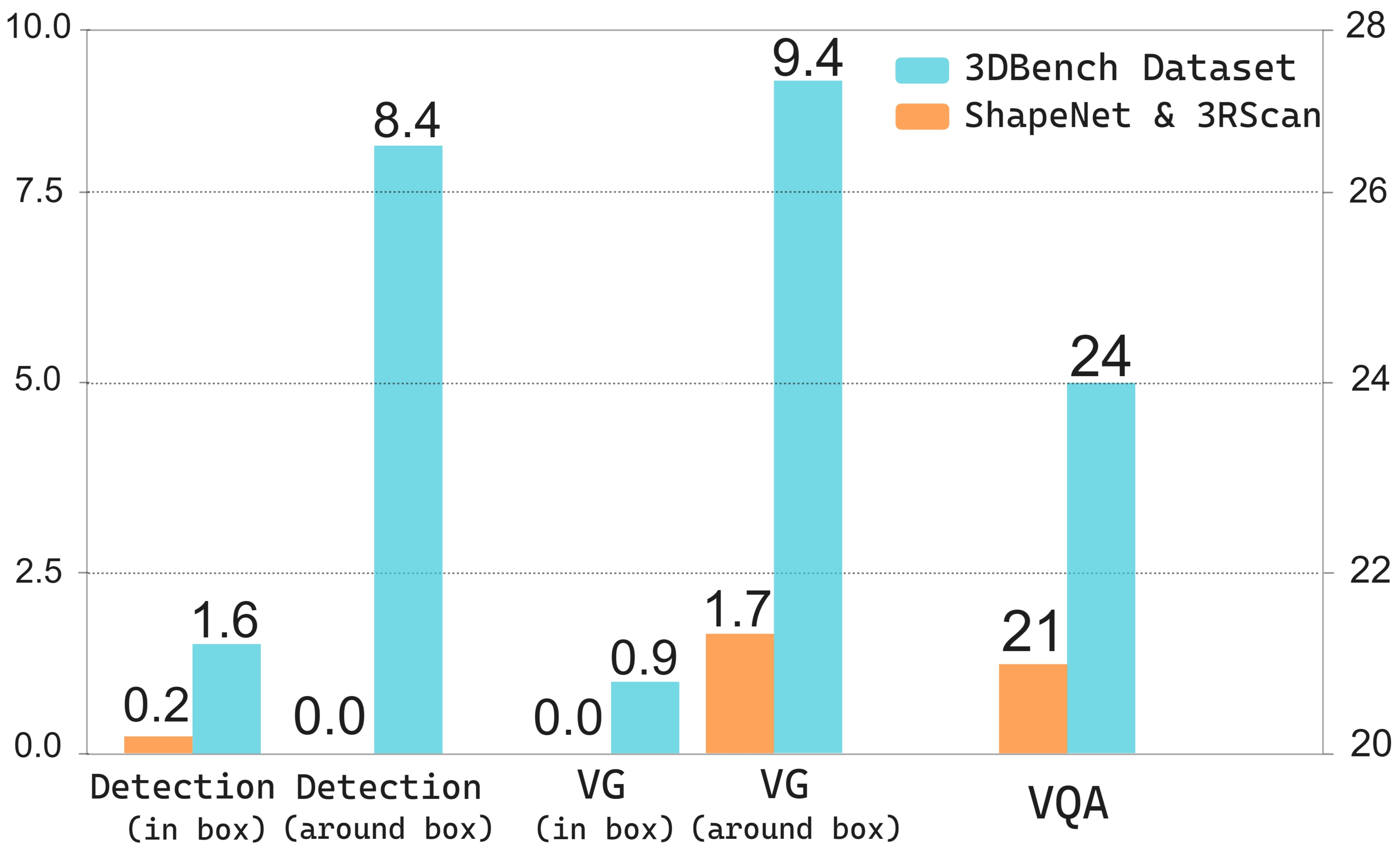} 	
	}

	\caption{Zero-shot results on cross-set validation of group E1 \& E2.}
	\label{fig:result1}
\end{figure}

Considering LAMM's questionable performance under conventional IOU evaluation, and subsequent corrections revealing a success rate below 1\%, we introduce ``in box" and ``around box" metrics as alternatives to the IOU metric, detailed in Section 3.3.
In Fig. \ref{fig:result1}, it is evident that models trained on our dataset outperformed those trained on ShapeNet and 3RScan in zero-shot results on the LAMM benchmark within the same training framework. This enhancement is attributed to the richer content and broader numerical distribution inherent in our dataset.

\subsection{Comparisons of Training Scales (E3)}
\begin{figure}[t]
	\centerline{
		\includegraphics[width=1\linewidth]{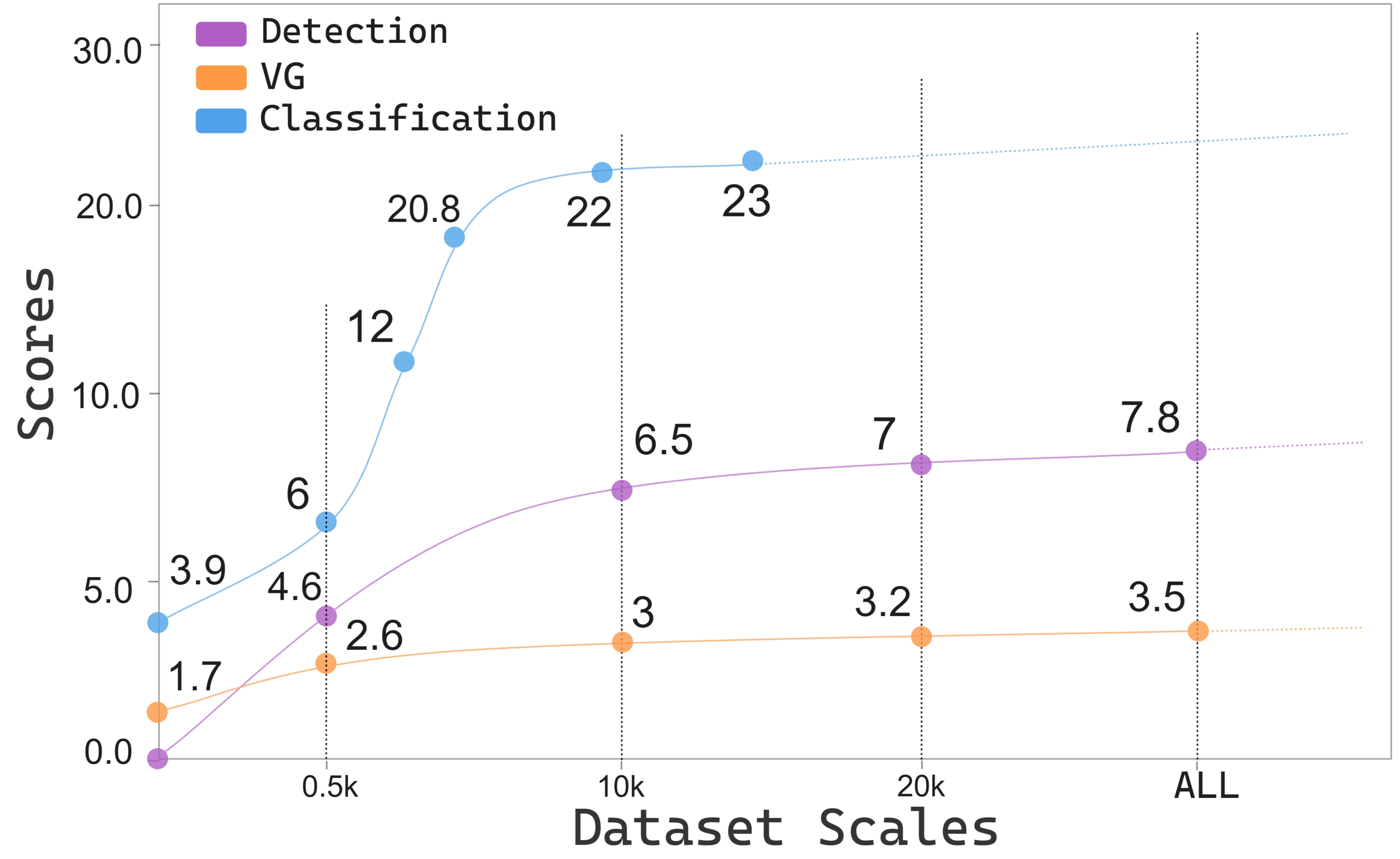} 	
	}

	\caption{Results on the impact of training set scale on 3D-LLMs performance.}
	\label{fig:result1-1}
\end{figure}

In Fig. \ref{fig:result1-1}, we present the performance of models re-trained with datasets of varying scales on 3DBench. This analysis serves to validate the effectiveness of dataset expansion and discern the maximum limit of performance enhancement with increased data volume. Initially, we assess the impact of dataset size by selecting 33k objects from the first 500 scenes, forming an object-level instruction-following dataset. Subsequently, we extract multiple objects from 30k scenes to create a scene-level dataset, re-training the large model with datasets of different sizes. We further explore the upper limit of performance improvement by expanding the object dataset, as illustrated by the solid line. The results reveal that, as the training dataset size increases, the performance of the LAMM model reaches a plateau around 20k. This suggests that, owing to the incapacity to learn additional features, the simple structure of an encoder plus LLM is insufficient for addressing multi-modal tasks in the 3D domain seamlessly.

% \begin{table}[htbp]
%     \centering
%     \begin{tabular}{lrr}
%         \toprule
%         Task\verb|\|Dataset  & 3DBench  & LAMM \\
%         \midrule
%         Detection(in box)     & 1.6          & 0.2       \\
%         Detection(around box)  & 8.4          & Failed      \\
%         Visual Grounding(in box) & 0.9          & Failed    \\
%         Visual Grounding(around box) & 9.4          & 1.7        \\
%         VQA                              & 24          & 21        \\
%         \bottomrule
%     \end{tabular}
%     \caption{Zero-shot results on 3DBench for the Baseline model trained on the original LAMM dataset and the model trained on the 3DBench dataset evaluated on the LAMM-Bench.}
%     \label{tab:booktabs}
% \end{table}

\subsection{Results after Re-training (E4)}

Table \ref{tab:results} displays the performance results of the LAMM on the 3DBench dataset before (to evaluate zero-shot ability) and after re-training, revealing a significant overall improvement in almost all tasks. Particularly notable is the approximately 20\% enhancement in classification and counting tasks, suggesting that the features of object point clouds and scene point clouds in the 3DBench dataset are readily learned by the large model. However, there is a discernible decline in performance for partial text generation and position relation tasks. This might be attributed to the use of GPT-3.5 for acquiring world knowledge, potentially causing a disparity in text quality compared to GPT-4, which is employed by LAMM. The experiments, considering comparisons with public datasets and the effectiveness of intrinsic features, collectively affirm the efficacy of the large-scale dataset obtained from our pipeline.

\begin{table}[t]
    \centering
    \resizebox {\linewidth} {!}{
    \begin{tabular}{c|ccc}
        \toprule
           & Re-training  & Zero-shot & $\Delta$  \\
        \midrule
        Detection     & 7.8          & Failed     & {\color{red} 7.8 $\uparrow$}  \\
        QA (scene)  & 59.5          & 62.5     & {\color{blue}3 $\downarrow$} \\
        Caption (scene) & 71.7          & 75.9   & {\color{blue}4.2 $\downarrow$}  \\
        QA (object)  & 81          & 66.5     & {\color{red} 14.5 $\uparrow$} \\
        Caption (object) & 89          & 70.2   & {\color{red} 18.8 $\uparrow$}  \\
        Classification  & 23          & 3.9    & {\color{red} 19.1 $\uparrow$} \\
        Visual Grounding     & 2.6          & 1.7  & {\color{red} 0.9 $\uparrow$} \\
        Counting     & 40          & 16.7   & {\color{red} 23.3 $\uparrow$}  \\
        Room Detection     & 7.4          & 3.3   & {\color{red} 4.1 $\uparrow$}   \\
        Position Relationship     & 25          & 31   & {\color{blue} 6 $\downarrow$}  \\
        Objections Relationship     & 73.5       & 63.1  & {\color{red} 10.4 $\uparrow$}  \\
        Navigation                   & 24.4       & 6.2    & {\color{red} 18.2$\uparrow$}  \\
        \bottomrule
    \end{tabular}}
    \caption{Results of LAMM on 3DBench, including both zero-shot test results and re-trained results.}
    \label{tab:results}
\end{table}

% \begin{table}
%     \centering
%     \begin{tabular}{lrrr}
%         \toprule
%           & Zero-shot& Small-scale  & Large-scale  \\
%         \midrule
%         Detection     & Failed          & 4.6      & 7.8     \\
%         VG  & 1.7          & 2.6     & 3.5    \\
%         Classification & 3.9          & 12     & 23    \\
%         \bottomrule
%     \end{tabular}
%     \caption{Results after fine-tuning.}
%     \label{tab:booktabs}
% \end{table}

\begin{figure*}[t]
	\centerline{
		\includegraphics[width=\linewidth]{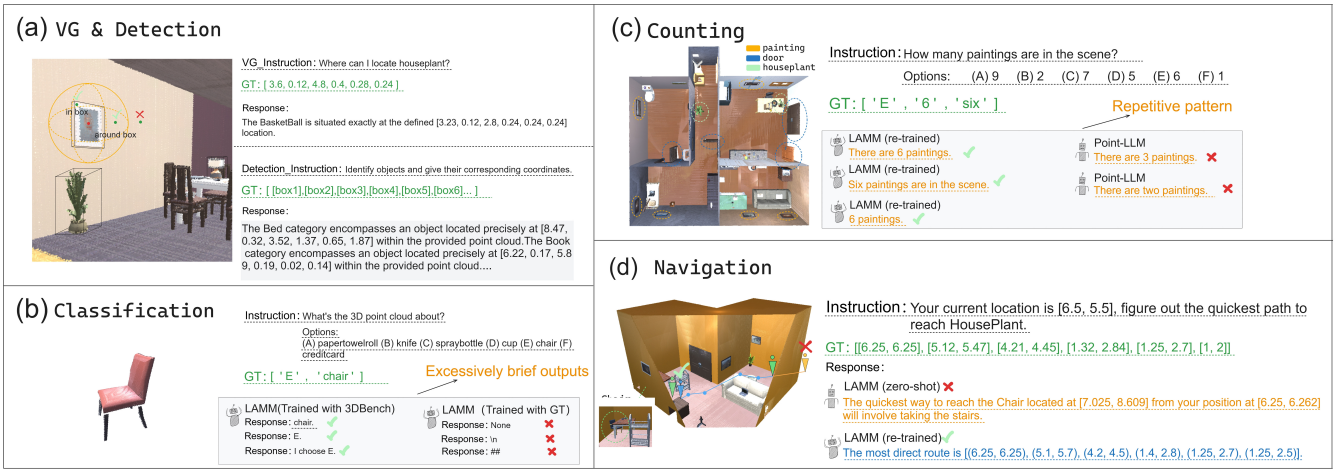} 	
	}
	\caption{The representative outcomes of 3D-LLMs are showcased as follows: (a) Visualization results for VG and detection tasks. (b) Constant answers in counting. (c) Null output in classification. (d) Divergent answers in navigation before and after re-training.
}
	\label{fig:insights}
\end{figure*}

\subsection{Comparisons of 3D-LLMs (E5)}

\begin{figure}[t]
	\centerline{
		\includegraphics[width=\linewidth]{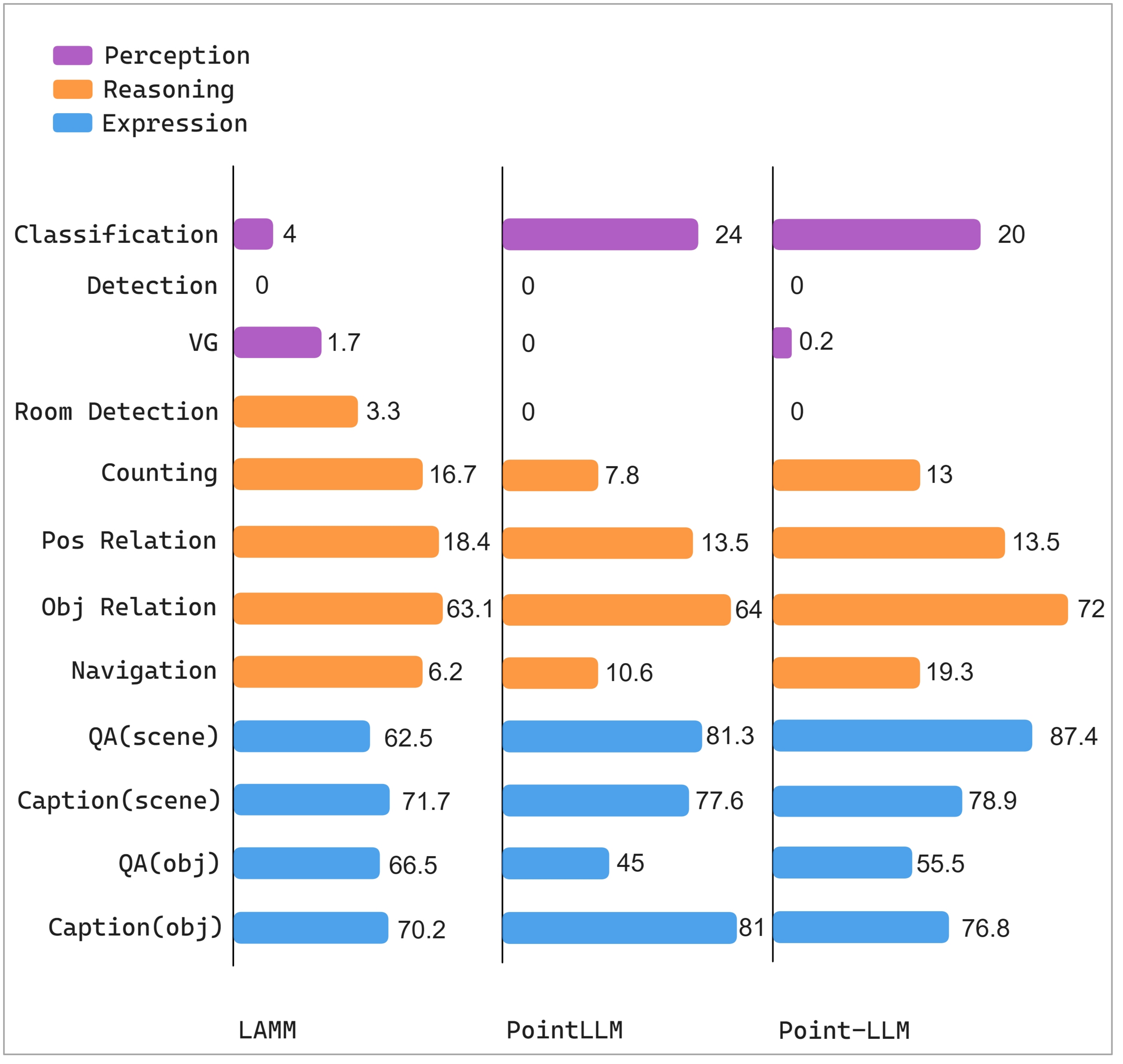} 	
	}

	\caption{Zero-shot evaluation results for existing 3D-LLMs.
}
	\label{fig:evaluationresults}
\end{figure}

In Fig. \ref{fig:evaluationresults}, PointLLM and Point-LLM exhibit significantly better performance in object-level tasks compared to LAMM. PointLLM benefits from its point cloud encoder utilizing point cloud color, and Point-LLM possesses a powerful multi-modal feature extractor, allowing them to gain more information compared to LAMM. However, in scene-level tasks, both models achieve inference results close to failure due to the training data only including objects. We plan to further assess their spatial understanding capabilities after re-training both 3D large models using the 3DBench dataset.

% \begin{table}[htbp]
%     \centering
%     \begin{tabular}{lrrr}
%         \toprule
%                           & LAMM  & PointLLM  & Point-LLM\\
%         \midrule
%         Classification     & 3.9         & 24     & 20  \\
%         Counting             & 16.7          & 7.8    &13     \\
%         QA                  & 62.5            & 81.3    & 87.4   \\
%         Caption                  & 71.7        & 77.6    & 78.9   \\
%         \bottomrule
%     \end{tabular}
%     \caption{ Zero-shot evaluation results for existing 3D large models.}
%     \label{tab:booktabs}
% \end{table}

% \begin{figure}[htbp]
% 	\centerline{
% 		\includegraphics[scale=0.1]{result3.png} 	
% 	}

% 	\caption{Zero-shot evaluation results for existing 3D large models.
% }
% 	\label{fig}
% \end{figure}

\section{Observations \& Analysis}

We summarize a series of observations to assess the performance of 3D-LLMs across diverse tasks, aiming to provide valuable insights to the academic community.

\textbf{Challenges in Incorporating Additional Features.} LAMM demonstrates limited zero-shot classification capabilities, yet a notable improvement is evident when incorporating color information from point clouds as in PointLLM and Point-LLM. Furthermore, we observe a general improvement in classification performance with an increased number of input point clouds, though it tends to reach a plateau rapidly. Additionally, the room detection task exhibits subpar performance for all models, highlighting ineffective feature extraction for substantial data volumes. We hypothesize that integrating more efficient feature extraction structures can enhance the performance of traditional vision tasks for 3D-LLMs.

% \textbf{Addressing Illusions in GPT Evaluations.} Comparative to GPT-4, employing datasets generated by GPT-3.5 yields lower text generation performance for LAMM. Nevertheless, unexpectedly high scores in Tab. \ref{tab:results} are observed in object-level captioning tasks. Upon examination, we identify model-induced illusions related to objects, resulting in high GPT scores. Modifying the prompts used in evaluation demonstrates that suitable prompts can mitigate the impact of illusions, producing reliable scores in scene-level evaluations. 

\textbf{Limitations in Spatial Understanding Capability.} 3D-LLMs originally demonstrate subpar performance in tasks involving positional relationships, VG, and object detection. This suggests that existing large models struggle to adeptly capture positional information within scene point clouds. By expanding the training set size and adopting more reasonable evaluation metrics, as depicted in Fig. \ref{fig:insights}(a), reveals a gradual alignment of inference results with the ground-truth, achieving accuracy several times higher than completely random outcomes. The spatial understanding capability of 3D-LLMs holds considerable potential for enhancement.

\textbf{Structure of Task Templates.} Training 3D-LLMs directly with ground-truth as responses may yield excessively brief outputs, resulting in empty inference results for certain tasks, as demonstrated in the case of classification in Fig. \ref{fig:insights}(b). 
Meanwhile, it is essential to vary the question-answer (QA) patterns within the templates. Upon examining the responses to Point-LLM counting tasks (zero-shot inference on 3DBench) in Fig. \ref{fig:insights}(c), we notice a repetitive pattern such as ``There are N objects" and ``The image shows N objects." Monotonous conversation templates could compromise the diversity and richness of outputs from 3D-LLMs. 
Therefore, it is advisable to employ GPT for generating predefined conversation templates and integrating ground-truth into these templates to construct an instruction-tuning dataset. 

\textbf{Navigation Benchmark.} Fig. \ref{fig:insights}(d) illustrates results for the navigation task. Navigation places heightened demands on the spatial perception and planning capabilities of 3D-LLMs, particularly in localization tasks. Our evaluation strategy yields improved scores for suboptimal output results, yet still leaves rooms for further improvements.

\section{Conclusion}

In this paper, we introduce 3DBench, a scalable benchmark designed for evaluating 3D-LLMs covering ten diverse multi-modal tasks with three types of evaluation metrics. Furthermore, we present a pipeline for automatically acquiring high-quality instruction-tuning datasets. Through extensive experiments, we validate the effectiveness of our dataset by cross-validate 3D-LLMs trained with various protocols. Our findings suggest that existing 3D-LLMs have considerable potential for further improvements in point cloud understanding and reasoning. We anticipate that our research will aid the research community in optimizing their models, and inspire the development of more efficient large models and high-quality instruction-tuning datasets.

\clearpage

%% The file named.bst is a bibliography style file for BibTeX 0.99c
\bibliographystyle{named}
\bibliography{ijcai24}

\end{document}